\documentclass[11pt,a4paper]{article}

\usepackage[utf8]{inputenc}
\usepackage[T1]{fontenc}
\usepackage{lmodern}
\usepackage[margin=1in]{geometry}
\usepackage{amsmath,amssymb}
\usepackage{booktabs}
\usepackage{graphicx}
\usepackage{tikz}
\usetikzlibrary{arrows.meta,positioning,calc}
\usepackage{caption}
\usepackage[colorlinks=true,linkcolor=blue!60!black,citecolor=blue!60!black,urlcolor=blue!60!black]{hyperref}
\usepackage{microtype}

\title{\textbf{ScoutRoute: Efficient Mixture-of-Experts Routing\\Using Internal Language Model Signals}}

\author{
  Delzaan Bharucha\thanks{delzaan123@gmail.com}\\
  \and
  Rajdeep Pandhare\thanks{Corresponding author: rajdeeppandhare@gmail.com}\\
  \\
  Department of Artificial Intelligence and Data Science\\
  Rajiv Gandhi Institute of Technology, Mumbai, Maharashtra, India
}
\date{}

\begin{document}
\maketitle

\begin{abstract}
Efficient expert routing is a fundamental challenge in Mixture-of-Experts (MoE) systems, particularly for heterogeneous collections of independently trained Large Language Models (LLMs). Existing routing approaches typically rely on jointly trained gating networks, prompt-based predictors, or computationally expensive multi-model inference, limiting their flexibility and efficiency.

This paper presents \emph{ScoutRoute}, a lightweight post-hoc routing framework that predicts expert competence using internal language model signals extracted from a single forward pass through an inexpensive scout model. Hidden-state representations are used to estimate cross-model competence through lightweight logistic probes, while logit-derived confidence measures are incorporated within a domain-aware routing strategy for efficient expert selection without modifying the underlying language models.

ScoutRoute is evaluated on five open-weight language models across four benchmark datasets covering mathematical reasoning, code generation, commonsense reasoning, and multidisciplinary knowledge. Experimental results show that ScoutRoute consistently outperforms individual routing signals while requiring only a single scout-model inference. The proposed competence probes also achieve superior calibration compared with conventional confidence measures, demonstrating that internal language model representations provide transferable information for accurate and efficient expert routing across heterogeneous language models.
\medskip

\end{abstract}

%------------------------------------------------------------
\section{Introduction}
Large Language Models (LLMs) have achieved remarkable performance across a wide range of tasks, including mathematical reasoning, code generation, question answering, and commonsense reasoning \cite{brown2020,chowdhery2022,touvron2023}. However, these improvements have been accompanied by rapidly increasing computational and memory requirements, making efficient inference a major challenge for real-world deployment.

Mixture-of-Experts (MoE) architectures address this challenge through conditional computation, activating only a subset of experts for each input rather than executing the entire model \cite{shazeer2017,fedus2022,mixtral2024}. While this significantly improves computational efficiency, effective expert routing remains a fundamental challenge. Conventional MoE routers are jointly trained with their experts and therefore cannot be directly applied to heterogeneous collections of independently trained language models \cite{fedus2022,mixtral2024}. Existing post-hoc routing methods instead rely on prompt embeddings, handcrafted confidence measures, or multi-model inference \cite{chen2023,hu2024,acuna2024}, often introducing additional computational overhead or failing to exploit information already available within the language model.

Recent studies have shown that internal language model representations contain rich information about both the input and the model's prediction confidence \cite{kadavath2022,alain2017}. Hidden-state embeddings capture semantic characteristics of the prompt, while logit-derived confidence measures provide useful estimates of prediction reliability \cite{kadavath2022}. These observations suggest that internal inference signals may be sufficient to predict the competence of candidate experts without executing every model.

Motivated by this observation, we propose ScoutRoute, a lightweight post-hoc routing framework for heterogeneous language model collections. ScoutRoute performs a single forward pass through an inexpensive scout model to extract hidden-state representations and logit-derived confidence measures. Hidden-state embeddings are used to estimate cross-model competence through lightweight logistic probes, while confidence statistics are incorporated within a domain-aware routing strategy to improve expert selection. Unlike conventional MoE routing methods, ScoutRoute requires no joint optimization with the underlying experts and performs routing using only a single scout-model inference.

ScoutRoute is evaluated on five open-weight language models across four benchmark datasets. Experimental results demonstrate that combining hidden-state competence estimation with logit-derived confidence improves routing accuracy while requiring only a single scout-model inference.

%------------------------------------------------------------
\section{Related Work}
\subsection{Routing in Mixture-of-Experts Systems}
Mixture-of-Experts architectures improve computational efficiency through conditional computation by activating only a subset of experts for each input \cite{shazeer2017}. Modern sparse architectures, including Switch Transformers and Mixtral, have demonstrated the effectiveness of this approach for scaling large language models \cite{fedus2022,mixtral2024}. However, these routing mechanisms rely on jointly trained gating networks and therefore cannot be directly applied to heterogeneous collections of independently trained language models. This limitation motivates lightweight post-hoc routing approaches that operate without joint optimization of routers and experts.

\subsection{Predictive Multi-Model Routing}
Recent work has explored predictive routing across heterogeneous language models to avoid evaluating every candidate expert. RouterBench introduced a standardized benchmark for evaluating routing strategies, while frameworks such as PROCEED and FrugalGPT estimate expert performance using prompt embeddings, external encoders, or query complexity to improve routing efficiency \cite{hu2024,acuna2024,chen2023}. Although effective, these approaches primarily rely on prompt-level representations or external embedding models rather than the internal representations generated during language model inference \cite{acuna2024,chen2023}. As a result, they cannot fully exploit the information already produced during a model's forward pass.

\subsection{Confidence Estimation and Hidden-State Probing}
Confidence estimation has long been used for selective prediction and expert routing \cite{guo2017,kadavath2022}. More recent studies have shown that hidden-state representations encode rich semantic and reasoning information that can be extracted using lightweight probes \cite{alain2017,kadavath2022}. At the same time, verbalized and logit-derived confidence estimates are often imperfectly calibrated, motivating approaches that leverage internal model representations for more reliable competence estimation. These findings suggest that hidden-state representations and logit-derived confidence provide complementary signals for expert selection.

Overall, existing routing approaches either depend on jointly trained routing networks or estimate expert competence using prompt-level representations and conventional confidence measures. ScoutRoute instead combines hidden-state competence estimation with logit-derived confidence extracted from a single scout-model inference, enabling lightweight post-hoc routing across heterogeneous language models without modifying or jointly training the underlying experts.

%------------------------------------------------------------
\section{Proposed Method}
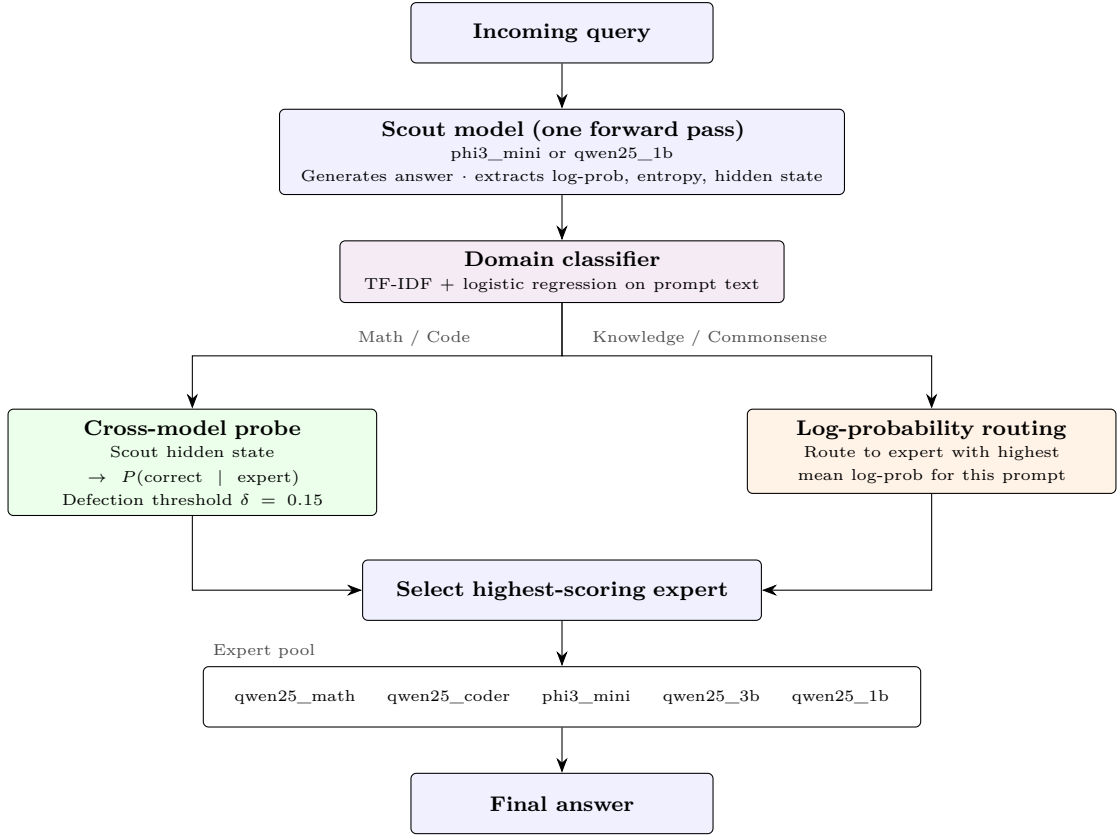
\begin{figure}[t]
\centering
\resizebox{0.92\textwidth}{!}{%
\begin{tikzpicture}[
  node distance=6mm,
  box/.style={draw, rounded corners=2pt, align=center, font=\scriptsize, inner sep=4pt, minimum height=8mm},
  wide/.style={box, text width=7.2cm},
  arr/.style={-{Stealth[length=2mm]}, thin},
  lab/.style={font=\tiny, text=black!70}
]
\node[box, fill=blue!6, minimum width=4cm] (q) {\textbf{Incoming query}};
\node[wide, fill=blue!6, below=of q] (scout) {\textbf{Scout model (one forward pass)}\\[-1pt]{\tiny phi3\_mini or qwen25\_1b}\\[-1pt]{\tiny Generates answer $\cdot$ extracts log-prob, entropy, hidden state}};
\node[box, fill=violet!8, below=of scout, text width=5.6cm] (dom) {\textbf{Domain classifier}\\[-1pt]{\tiny TF-IDF + logistic regression on prompt text}};
\node[box, fill=green!8, below left=14mm and -5mm of dom, text width=4.6cm] (probe) {\textbf{Cross-model probe}\\[-1pt]{\tiny Scout hidden state $\rightarrow P(\text{correct}\mid\text{expert})$}\\[-1pt]{\tiny Defection threshold $\delta=0.15$}};
\node[box, fill=orange!10, below right=14mm and -5mm of dom, text width=4.6cm] (lp) {\textbf{Log-probability routing}\\[-1pt]{\tiny Route to expert with highest}\\[-1pt]{\tiny mean log-prob for this prompt}};
\node[box, fill=blue!6, below=34mm of dom, text width=5.0cm] (sel) {\textbf{Select highest-scoring expert}};
\node[box, below=of sel, minimum width=9.5cm, inner sep=5pt] (pool) {%
 \tiny\begin{tabular}{@{}ccccc@{}}
 qwen25\_math & qwen25\_coder & phi3\_mini & qwen25\_3b & qwen25\_1b
 \end{tabular}};
\node[lab, above=-1pt of pool.north west, anchor=south west] {Expert pool};
\node[box, fill=blue!6, below=of pool, minimum width=4cm] (ans) {\textbf{Final answer}};

\draw[arr] (q) -- (scout);
\draw[arr] (scout) -- (dom);
\draw[arr] (dom.south) -- ++(0,-7mm) -| node[lab, pos=0.2, above]{Math / Code} (probe.north);
\draw[arr] (dom.south) -- ++(0,-7mm) -| node[lab, pos=0.2, above]{Knowledge / Commonsense} (lp.north);
\draw[arr] (probe.south) |- (sel.west);
\draw[arr] (lp.south) |- (sel.east);
\draw[arr] (sel) -- (pool);
\draw[arr] (pool) -- (ans);
\end{tikzpicture}}
\caption{Overall architecture of the proposed ScoutRoute framework. The figure illustrates the single-pass routing pipeline, including scout model inference, internal signal extraction, competence estimation, domain-aware routing, and expert selection.}
\label{fig:arch}
\end{figure}

ScoutRoute is a lightweight post-hoc routing framework that performs expert selection using internal language model signals extracted from a single scout-model inference. The framework combines hidden-state competence estimation with logit-derived confidence to predict the most suitable expert without executing every candidate model. Figure~\ref{fig:arch} illustrates the overall routing pipeline.

\subsection{Problem Formulation}
Consider a collection of $N$ independently trained language models, where each expert exhibits different capabilities across reasoning domains. Given an input query $q$, the objective is to identify the expert expected to generate the highest-quality response while minimizing inference cost.

Conventional routing approaches evaluate every candidate expert before selecting the best response, resulting in computational cost that increases linearly with the number of available experts. ScoutRoute instead seeks to predict the most suitable expert directly from a single scout-model inference.

The routing objective is formulated as learning a routing function
\begin{equation}
R(q): q \rightarrow E_i,
\end{equation}
which predicts the most appropriate expert without executing every candidate model.

ScoutRoute approaches this problem by predicting expert competence from internal language model signals extracted during a single forward pass through a lightweight scout model. Rather than evaluating every expert directly, the framework estimates the probability that each expert will correctly answer the query before inference.

\subsection{Competence Estimation}
ScoutRoute estimates the competence of each candidate expert directly from the hidden-state representation produced by the scout model. This design is inspired by prior work showing that internal language model representations encode predictive information about model behaviour \cite{alain2017,kadavath2022}, but extends these ideas to cross-model expert competence estimation for heterogeneous routing. Rather than executing every expert, lightweight competence probes predict the probability that each expert will correctly answer the input query.

Before probe training, the hidden-state embeddings are projected into a lower-dimensional space using Principal Component Analysis (PCA) to reduce feature redundancy and improve generalization \cite{jolliffe2002}. An independent logistic regression classifier is then trained for each candidate expert. The predicted competence of expert $E_i$ is given by
\begin{equation}
P_i = \sigma\!\left(\mathbf{w}_i^{\top}\tilde{\mathbf{e}} + b_i\right),
\end{equation}
where $\tilde{\mathbf{e}}$ denotes the PCA-transformed embedding, $\sigma(\cdot)$ is the sigmoid activation function, and $P_i\in[0,1]$ represents the predicted probability that expert $E_i$ will produce a correct response.

The competence probes are trained using benchmark-specific correctness labels while keeping all language models frozen, allowing ScoutRoute to operate as a lightweight post-hoc routing framework without requiring joint optimization of the experts. The resulting competence scores serve as the primary routing signal and are subsequently combined with logit-derived confidence measures using the domain-aware routing strategy described in the next subsection.

\subsection{Domain-Aware Routing}
ScoutRoute combines hidden-state competence estimation with logit-derived confidence through a domain-aware routing strategy. Competence probes are used for specialist reasoning tasks (GSM8K and HumanEval), while mean log probability is used for broader reasoning tasks (CommonsenseQA and MMLU).

The query domain is first predicted using a TF--IDF representation \cite{salton1988} and a logistic regression classifier, enabling domain-aware routing with minimal computational overhead. The routing score assigned to expert $E_i$ is
\begin{equation}
S_i(q)=
\begin{cases}
P_i, & D(q)\in\{\text{GSM8K},\text{HumanEval}\},\\[2pt]
C_{\log}, & D(q)\in\{\text{CommonsenseQA},\text{MMLU}\},
\end{cases}
\end{equation}
where $P_i$ is the probe-predicted competence score and $C_{\log}$ is the scout model's mean log probability.

For specialist domains, routing switches away from the designated expert only when
\begin{equation}
P_g - P_s > \delta,
\end{equation}
where $P_s$ is the specialist competence score, $P_g$ is the highest competing score, and $\delta$ is an empirically selected threshold. The selected expert is
\begin{equation}
E^{*}=\arg\max_i S_i(q),
\end{equation}
and is the only model executed during inference. An oracle-domain variant, using ground-truth domain labels, is also evaluated as an upper performance bound.

%------------------------------------------------------------
\section{Experimental Setup}
\subsection{Benchmark Datasets}
Experiments were conducted on four benchmark datasets spanning mathematical reasoning, code generation, commonsense reasoning, and multidisciplinary knowledge: GSM8K \cite{cobbe2021}, HumanEval \cite{chen2021}, CommonsenseQA \cite{talmor2019}, and MMLU \cite{hendrycks2021}. These benchmarks provide diverse reasoning tasks for evaluating expert routing across heterogeneous language models.

\subsection{Candidate Language Models}
Experiments were conducted using five open-weight language models representing both specialist and general-purpose capabilities. The evaluated models include DeepSeek-R1-Distill-Llama-8B \cite{deepseek2025}, Llama-3.1-8B-Instruct \cite{dubey2024}, Qwen2.5-7B-Instruct, Qwen2.5-Coder-7B-Instruct, and Qwen2.5-Math-7B-Instruct \cite{yang2024}. The selected models exhibit complementary strengths across the benchmark datasets, making them well suited for evaluating expert routing in heterogeneous language model collections.

\subsection{Probe Training}
To evaluate probe-based competence estimation, the benchmark prompts were divided into disjoint training and evaluation sets. Following the experimental protocol, 30\% of the prompts from each dataset were randomly selected for probe training, while the remaining 70\% were reserved exclusively for routing evaluation. The split was performed independently within each benchmark to ensure that all reasoning domains were represented in both partitions.

For each candidate expert, hidden-state embeddings extracted from the scout model were first projected into a lower-dimensional space using PCA \cite{jolliffe2002}. Logistic regression classifiers with balanced class weights were then trained to predict whether the corresponding expert would correctly answer a given query. Balanced weighting mitigated the effect of class imbalance arising from differences in expert performance across datasets. Probe performance was validated using an internal 80/20 split of the training partition before retraining on the full training set for final evaluation.

\subsection{Implementation and Evaluation}
All experiments were implemented in Python using the Hugging Face Transformers library \cite{wolf2020} together with scikit-learn \cite{pedregosa2011}. Principal Component Analysis was used for dimensionality reduction, while logistic regression served as the competence estimation model. Hidden-state representations and logit-derived confidence measures were extracted during a single forward pass through each scout model.

The proposed framework was evaluated using identical train--evaluation splits across all routing strategies to ensure fair comparison. Random seeds were fixed throughout the experiments to ensure reproducibility.

The evaluation metrics were routing accuracy, probe discrimination, and calibration quality. Routing performance was evaluated using routing accuracy, defined as the proportion of evaluation prompts for which the selected expert produced the correct answer. Probe quality was measured using the Area Under the Receiver Operating Characteristic Curve (AUROC), while calibration quality was assessed using the Expected Calibration Error (ECE) \cite{guo2017}. Lower ECE values indicate better agreement between predicted confidence and empirical correctness. For comparison, random expert selection and an oracle router were included as lower and upper performance bounds, respectively.

%------------------------------------------------------------
\section{Experimental Results}
This section evaluates ScoutRoute using routing accuracy, probe quality, calibration, and scout model analysis across multiple benchmark datasets and heterogeneous language models.

\subsection{Overall Routing Performance}
We first compare ScoutRoute with standalone routing strategies, including random routing, mean log probability, negative entropy, probe-based competence estimation, and an oracle router representing the theoretical upper bound. Table~\ref{tab:standalone} summarizes the routing accuracy of the standalone methods.

\begin{table}[h]
\centering
\caption{Routing accuracy of standalone routing strategies.}
\label{tab:standalone}
\begin{tabular}{lc}
\toprule
Routing Strategy & Accuracy \\
\midrule
Random Routing & 35.36\% \\
Negative Entropy & 63.47\% \\
Probe Score & 63.16\% \\
Mean Log Probability & 65.02\% \\
Oracle Router & 82.66\% \\
\bottomrule
\end{tabular}
\end{table}

All learned routing strategies substantially outperform random routing, confirming that internal language model signals provide effective routing cues. Among the standalone methods, mean log probability achieves the highest accuracy (65.02\%), followed closely by the competence probes (63.16\%). The oracle router establishes an upper performance bound of 82.66\%. Since no individual routing signal consistently performs best across all reasoning domains, ScoutRoute combines competence estimation and logit-derived confidence through the proposed domain-aware routing strategy, evaluated in the following subsection.

\subsection{Domain-Aware ScoutRoute}
Table~\ref{tab:scoutroute} compares the proposed ScoutRoute framework with the standalone routing strategies.

\begin{table}[h]
\centering
\caption{Comparison of standalone routing signals and the proposed ScoutRoute framework.}
\label{tab:scoutroute}
\begin{tabular}{lc}
\toprule
Routing Method & Accuracy \\
\midrule
Random Routing & 35.54\% \\
Negative Entropy & 63.47\% \\
Probe Score & 63.16\% \\
Mean Log Probability & 65.02\% \\
\textbf{ScoutRoute (Domain-Aware)} & \textbf{67.80\%} \\
Oracle Router & 82.66\% \\
\bottomrule
\end{tabular}
\end{table}

ScoutRoute achieves the highest routing accuracy among all practical routing strategies, improving from 65.02\% with the strongest standalone signal (mean log probability) to 67.80\%. This improvement demonstrates that hidden-state competence estimation and logit-derived confidence provide complementary routing information. Although an oracle router establishes an upper bound of 82.66\%, ScoutRoute offers substantially better routing performance than individual signals while requiring only a single scout-model inference.

\subsection{Probe Quality Analysis}
To evaluate the competence probes independently of the routing strategy, we measure the Area Under the Receiver Operating Characteristic Curve (AUROC) for each scout model. Table~\ref{tab:auroc} summarizes the results.

\begin{table}[h]
\centering
\caption{Competence probe performance measured using AUROC.}
\label{tab:auroc}
\begin{tabular}{lc}
\toprule
Scout Model & AUROC \\
\midrule
Qwen2.5-1B & 0.8957 \\
Phi-3 Mini & 0.8610 \\
Qwen2.5-3B & 0.8231 \\
Qwen2.5-Math & 0.7240 \\
Qwen2.5-Coder & 0.6823 \\
\bottomrule
\end{tabular}
\end{table}

The competence probes achieve strong discrimination performance across all evaluated scout models, with AUROC values ranging from 0.6823 to 0.8957. Qwen2.5-1B produces the highest AUROC, followed by Phi-3 Mini and Qwen2.5-3B, indicating that compact general-purpose models provide highly transferable hidden-state representations for competence prediction. In contrast, the specialist models Qwen2.5-Math and Qwen2.5-Coder achieve lower AUROC values, suggesting that their representations are less effective for predicting the performance of heterogeneous experts.

\subsection{Calibration Analysis}
To assess the reliability of the routing signals, we evaluate their calibration using the Expected Calibration Error (ECE). Table~\ref{tab:ece} summarizes the calibration performance across all scout models.

\begin{table}[h]
\centering
\caption{Expected Calibration Error (ECE) of different routing signals. Lower values indicate better calibration.}
\label{tab:ece}
\begin{tabular}{lccc}
\toprule
Scout Model & Log Prob. & Neg. Entropy & Probe \\
\midrule
Qwen2.5-1B & 0.4691 & 0.4299 & \textbf{0.1456} \\
Qwen2.5-3B & 0.4570 & 0.4548 & \textbf{0.1289} \\
Phi-3 Mini & 0.4723 & 0.3155 & \textbf{0.1313} \\
Qwen2.5-Math & 0.4155 & 0.3236 & \textbf{0.2313} \\
Qwen2.5-Coder & 0.3632 & 0.3868 & \textbf{0.2433} \\
\bottomrule
\end{tabular}
\end{table}

Across all evaluated scout models, the competence probes consistently achieve the lowest ECE, indicating substantially better calibration than both mean log probability and negative entropy. This improved calibration suggests that probe-based competence estimates provide more reliable confidence for expert routing than logit-derived confidence measures alone.

\subsection{Scout Model Comparison}
To evaluate the effect of the scout model on routing performance, each candidate language model was evaluated as the scout while keeping the remaining routing pipeline unchanged. Table~\ref{tab:scouts} summarizes the resulting routing accuracy.

\begin{table}[h]
\centering
\caption{Routing accuracy obtained using different scout models.}
\label{tab:scouts}
\begin{tabular}{lcc}
\toprule
Scout Model & Probe Router & ScoutRoute \\
\midrule
Qwen2.5-1B & 65.33\% & 67.80\% \\
Qwen2.5-3B & 62.85\% & 66.87\% \\
Phi-3 Mini & 65.33\% & \textbf{68.42\%} \\
Qwen2.5-Math & 58.51\% & 65.33\% \\
Qwen2.5-Coder & 61.92\% & 67.18\% \\
\bottomrule
\end{tabular}
\end{table}

ScoutRoute consistently improves routing accuracy across all scout models. Phi-3 Mini achieves the highest routing accuracy (68.42\%), followed closely by Qwen2.5-1B, while the specialist models achieve comparatively lower performance. The smaller variation observed for ScoutRoute compared with the probe-only router indicates that the domain-aware routing strategy is more robust to the choice of scout model.

\subsection{Specialist Routing Analysis}
To evaluate ScoutRoute on specialist tasks, we analyse routing decisions for GSM8K and HumanEval, which are associated with the mathematical and code-generation specialists, respectively. Table~\ref{tab:specialist} summarizes the routing behaviour.

\begin{table}[h]
\centering
\caption{Analysis of specialist routing behaviour.}
\label{tab:specialist}
\begin{tabular}{llccc}
\toprule
Domain & Specialist & Routed (\%) & Specialist Acc. & Defection Acc. \\
\midrule
GSM8K & Qwen2.5-Math & 72.86 & 86.27 & 73.68 \\
HumanEval & Qwen2.5-Coder & 72.17 & 67.47 & 15.62 \\
\bottomrule
\end{tabular}
\end{table}

ScoutRoute selected the designated specialist for approximately 72\% of queries across both specialist benchmarks. While the mathematics specialist remained highly effective on GSM8K (86.27\% accuracy), defections from the HumanEval specialist resulted in substantially lower accuracy (15.62\%), supporting the use of the proposed specialist defection threshold.

%------------------------------------------------------------
\section{Discussion}
The experimental results demonstrate that internal language model signals provide an effective basis for post-hoc expert routing across heterogeneous language models. While competence probes and logit-derived confidence measures each provide competitive routing performance, neither consistently performs best across all reasoning domains. By combining these complementary signals through a domain-aware routing strategy, ScoutRoute achieves higher routing accuracy than either signal alone while requiring only a single scout-model inference.

An important finding is that compact general-purpose language models serve as more effective scout models than larger specialist models. Although Qwen2.5-1B achieved the highest probe discrimination performance, Phi-3 Mini produced the strongest end-to-end routing accuracy, indicating that effective scout models need not be the largest available. These results suggest that broadly trained models learn internal representations that transfer well across heterogeneous expert collections. The calibration analysis further demonstrates that competence probes provide substantially more reliable confidence estimates than logit-derived confidence measures.

Overall, the findings support the central hypothesis of this work: internal language model representations contain transferable information that can accurately predict the competence of other language models. Consequently, ScoutRoute provides an efficient and practical routing framework that avoids jointly trained routing networks and exhaustive multi-model inference while maintaining competitive routing performance.

%------------------------------------------------------------
\section{Conclusion}
This paper introduced ScoutRoute, a lightweight post-hoc routing framework for efficient expert selection across heterogeneous collections of large language models. By combining hidden-state competence probes with logit-derived confidence measures through a domain-aware routing strategy, ScoutRoute performs expert selection using only a single scout-model inference without requiring jointly trained routing networks.

Experiments demonstrate that ScoutRoute consistently improves routing accuracy while requiring only a single scout-model inference. The results further indicate that compact general-purpose language models provide effective scout representations for cross-model competence estimation. Future work will investigate extending ScoutRoute to larger expert collections, additional reasoning domains, improved domain classification, online adaptation of competence probes, and multimodal foundation models.

%------------------------------------------------------------

\end{document}